**Pl@ntNet machine-to-machine identification service with Pl@ntNet API: enabling automated biodiversity monitoring through large-scale plant identification**

Benjamin Bourel[1], Jean-Christophe Lombardo[1], Alexis Joly[1], Mathias Chouet[2], Thomas Paillot[1], Antoine Affouard[1] , Vanessa Hequet[2] , Hervé Goëau[2], Pierre Bonnet[2].

[1]Inria, LIRMM, Montpellier, France

[2]AMAP, Univ Montpellier, CIRAD, CNRS, INRAE, IRD, Montpellier, France

**Abstract**

**1.** Monitoring plant biodiversity requires identifying large numbers of plants across space and time. Although artificial intelligence (AI) can now identify plant species from images with high accuracy, most existing systems rely on users manually submitting images. As ecological monitoring increasingly generates large image collections from cameras, drones, robots and online platforms, there is a growing need for plant identification services that can be directly integrated into automated data-processing workflows.

**2.** Here, we present Pl@ntNet's machine-to-machine AI plant identification service. This service uses an application programming interface (API) to provide programmatic access to Pl@ntNet's plant recognition models and the associated taxonomic processing. This service enables the implementation of automated identification workflows without human intervention and can thus directly link large image collections and automated observation systems to an AI-based plant identification system.

**3.** We establish that this service, which covers 85,000 species from around the world, (1) has a species prediction accuracy of 0.89, (2) process images up to 666 times faster than workflows requiring human intervention, (3) and has processed over 84 million identification requests since its launch, including more than 40 million in 2025 alone. We demonstrate how it has already enabled the implementation of a new generation of ecological workflows: identifying plants from images of insects to reconstruct the plant-pollinator interactions, analysing geolocated images from social media to study plant distribution, detecting invasive plants using images taken at the roadside, and identifying weeds using agricultural robots. Finally, we illustrate its application using a reproducible example: identifying plants from insect photos on iNaturalist. In approximately 30 min on a standard laptop, 2,424 images were processed to generate a dataset of 708 plant-pollinator interactions, with a weighted precision of 0.98 for flower detection and an accuracy of 0.88 for genus plant identification.

**4.** In conclusion, this machine-to-machine species identification is already operating on a considerable scale and meets a genuine need. All of this paves the way for a new paradigm in biodiversity monitoring, in which automated sensors and image collections can be directly connected to AI-based identification services.

## 1. Introduction

The global decline in plant biodiversity is a major challenge and requires large-scale monitoring. However, the shortage of expert taxonomists limits the identification of plant species and, consequently, large-scale biodiversity monitoring. To overcome this limitation, citizen science programs have emerged as a powerful solution to increase observation effort (Dickinson et al., 2010). In parallel, advances in artificial intelligence (AI) have made it possible to automatically identify plant species from images with accuracy approaching expert-level performance (Bonnet et al., 2018; Wäldchen & Mäder, 2018). These developments have led to the emergence of large-scale participatory platforms such as Pl@ntNet and iNaturalist, which combine citizen observations with automated identification to support biodiversity research, conservation, and environmental monitoring (Bonnet et al., 2020). These AI-based plant identification platforms still rely primarily on users manually submitting images (Joly et al., 2016). Despite its effectiveness in citizen science, this paradigm limits the integration of automated identification into ecological data streams derived from large-scale automated observation systems such as agricultural monitoring systems or drones. As biodiversity monitoring increasingly relies on these digital data sources, the ability to perform machine-to-machine species identification is becoming a key requirement for next-generation ecological infrastructures. To enable this transition, we must move from standalone identification applications to interoperable services that can be integrated into automated workflows.

It is within this context that we have developed the Pl@ntNet machine-to-machine service. This service is based on an application programming interface (API) that provides direct access to the Pl@ntNet plant recognition system. By enabling automated queries without human intervention, this service allows AI-based plant identification to be integrated into ecological data flows. This machine-to-machine identification service provides access to the entire identification workflow of Pl@ntNet, including taxonomic post-processing. This represents a significant paradigm shift for AI-based biodiversity monitoring as, to our knowledge, there is currently no equivalent on other similar platforms of this scale. Since its launch in 2018, the Pl@ntNet machine-to-machine service has enabled a wide range of applications, such as the automatic detection of plants using images from social media, the reconstruction of ecological interactions using photographs from citizen science, or the monitoring of the spread of invasive plants (August et al., 2020; Goëau et al., 2026; Pernat et al., 2024). In 2025, it processed an average of 115 000 requests every day. This demonstrates the importance and necessity of the paradigm shift we are proposing through this service. Here, we present the Pl@ntNet machine-to-machine service and describe how providing large-scale, programmatic access to plant identification models enables new forms of biodiversity observation. We detail the design of the interface, its current usage, several real-world applications, and discuss how such services contribute to the development of interoperable monitoring systems aligned with next-generation ecological observatories.

## 2. Pl@ntNet and Pl@ntNet machine-to-machine service

### *2.1. A brief introduction to Pl@ntNet*

Pl@ntNet has established itself as a leading citizen initiative in the field of plant identification using AI (Lefort et al., 2025). Pl@ntNet's AI identification models are currently based on vision transformers. At the time of writing, the Pl@ntNet web and mobile platforms have processed 1.4 billion queries, including 221 million in 2025. A query corresponds to the submission of a raw observation of one or more images of the same plant. These raw observations are processed in real time by Pl@ntNet's AI models to return a list of species to users, ranked in descending order of probability. The raw observations can then be converted into shared observations by the 9 million registered users. The shared observation includes the raw observation associated with metadata (*e.g.*, date, location, type of organ when specified) and an initial species label proposed either by the AI model or manually selected by the user. These shared observations are incorporated into the Pl@ntNet collaborative database, which currently contains 30 million observations shared publicly by their authors and associated with one of the 85 000 species covered. Each shared observation can then receive additional annotations in the form of votes from other users. Users can confirm the proposed species or suggest another identification. Pl@ntNet does not rely on simple majority voting, but implements a cooperative tag aggregation strategy that estimates user expertise through an iterative trust rating mechanism (Lefort et al., 2025). Shared observations that achieve sufficient consensus and confidence thresholds are marked as valid and incorporated into the database used to regularly train Pl@ntNet models. Thanks to all this, Pl@ntNet models are currently regularly trained and tested using 18 million plant images annotated with curated labels.

Pl@ntNet's performance is regularly assessed using a test dataset randomly sampled from the 18 million images annotated with curated labels. In March 2026, the accuracy of the global flora model was 0.886 for top-1 and 0.901 for top-5 on single images (**1**). Performance is also evaluated on two datasets external to Pl@ntNet, which have been published and are freely accessible: Swiss Plant dataset (Popp, Provost, et al., 2025; Popp, Zimmermann, et al., 2025) and the iNaturalist 2021 validation dataset (Horn et al., 2021). After alignment with Pl@ntNet's taxonomic reference at species level, Swiss Plant comprises 5,619 records of 561 species in Switzerland (**Table S1**). For iNaturalist 2021, only those observations of Tracheophyta that could be aligned were included, amounting to 37,160 observations of 3,716 species worldwide (**Table S2**). The accuracy of the global flora model is 0.864 for top-1 and 0.963 for top-5 for Swiss Plant, and 0.721 for top-1 and 0.876 for top-5 for iNaturalist 2021, again using a single image. Performance was relatively poor with iNaturalist 2021 because, unlike the other datasets, the image size is smaller than the 800×800 pixels recommended by Pl@ntNet: 500×350 pixels on average. Finally, it should be noted that for Swiss Plant, the accuracy rises to 0.879 for top-1 and 0.973 for top-5 for the Central European flora model. This demonstrates the value of having models based on the POWO floras (Powo, 2026) within the Pl@ntNet automated identification service.

| Testset | Pl@ntNet model | Mean image size | Top1 | Top5 |
|---|---|---|---|---|
| Pl@ntNet testset | world-flora | > 800×800 pixels | 0.886 | 0.901 |
| iNaturalist 2021 valset | world-flora | 500×350 pixels | 0.721 | 0.876 |
| Swiss Plant dataset | world-flora | 1100×1200 pixels | 0.864 | 0.963 |
| Swiss Plant dataset | middle-europe | 1100×1200 pixels | 0.879 | 0.973 |

**Figure 1 :** Trends in the use of the Pl@ntNet API for machine-to-machine identification

### 2.2. Main functionality of Pl@ntNet machine-to-machine service

The Pl@ntNet machine-to-machine service is accessed via the Pl@ntNet API. The Pl@ntNet API enables users to interact with the Pl@ntNet service (Pl@ntNet AI models and associated pre- and post-processing) via HTTPS (HyperText Transfer Protocol Secure) requests that can be automated using Python, R, Java and JavaScript scripts. The central element of the Pl@ntNet machine-to-machine service is the "/identify" endpoint of the Pl@ntNet API. Here, '/identify' allows the user to identify a plant species from one or more images. The request accepts up to five images representing the same plant. Optional parameters allow the user to choose the AI model to use, specify the plant's native flora (*e.g.*, to optimise identification), indicate the observed organ (*e.g.*, leaf, flower, fruit, bark), limit the number of results (*e.g.*, the size of the list of most likely species) or obtain detailed information such as results by genus or family. The response to the query is returned in JSON format.

The response for "/identify" query is: **(1)** the scientific name based on POWO (Powo, 2026) and common name of the most likely species; **(2)** the list of most likely species ranked in descending order of confidence score; **(3)** the confidence score for each species; **(4)** the most similar image in the database for each species **(5)**; identifiers for GBIF (*GBIF*, 2026), POWO (*Powo*, 2026) and IUCN (*IUCN*, 2026) for each species. When the corresponding options are enabled, the response also indicate: **(6)** the automatically detected organ for each image with its confidence score; **(7)** additional predictions at higher taxonomic levels, including genus and family, each with a score, taxonomic information, external identifiers, and the associated illustrative images. **(8)** Finally, the response will include the list of parameters used for the query and the status of the user's identification quota.

### 2.3. Access and user-friendliness of Pl@ntNet machine-to-machine service

Access to the API and the machine-to-machine service requires authentication via an API key, which enables quotas and throughput limits to be applied to ensure fair use. An API key can be obtained by creating an account on the Pl@ntNet API website (*My Pl@ntNet API*, 2026). Three options are available: a free version limited to 500 identifications per day, a free option for education and research with customisable quotas available on request, and a paid option. Comprehensive documentation on the Pl@ntNet API website facilitates the integration of the service into ecological and biodiversity monitoring workflows, with instructions on account creation, image preparation and identification queries, as well as reproducible examples in Python, R, Java and JavaScript, and a description of inputs, outputs and confidence scores.

### 2.4. Best practices for ecological workflows

To maximize the reliability of the machine-to-machine identification service in ecological pipelines, users should optimize two main parameters: image quality and confidence score thresholds. First, image resolution directly impacts taxonomic accuracy. As demonstrated by the lower performance on the iNaturalist 2021 validation set (**Table 1**), submitting images below the recommended 800x800 pixels significantly degrades the model's predictive capabilities. Second, users must calibrate the confidence score threshold according to their specific experimental design. The API returns a confidence score for each predicted species. Setting a high threshold (e.g., > 0.8) maximizes accuracy by minimizing false positives, which is critical for strict biodiversity inventories. Conversely, setting a lower threshold, such as the 0.5 value applied in the plant-pollinator workflow in Section 4, favors recall. This is more appropriate for broad exploratory analyses where missing a true positive is more detrimental than processing false positives.

## 3. The impacts of Pl@ntNet machine-to-machine service

### *3.1. Popular support and effectiveness*

Since its launch in 2019, the Pl@ntNet machine-to-machine service has experienced exponential growth. In 2025, more than 7,000 new accounts brought the total number to 16,250 (**Figure 1**), whilst more than 40 million queries were made, bringing the total to 84 million. With the number of queries doubling every year, these figures demonstrate the growing adoption of the machine-to-machine species identification service for next-generation ecological monitoring. At the same time, improvements to the infrastructure and code reduced the average identification time from 1.3 to 0.9 seconds between 2024 and 2025, despite the doubling of queries. With up to 20 simultaneous requests per account, 1,000 images can be processed in around 45 seconds, compared with 8.3 hours manually via the web interface (~30 s/image), representing a processing speed 666 times faster and requiring no human intervention. Finally, between 2024 and 2025, only 4.5 per cent of requests failed.

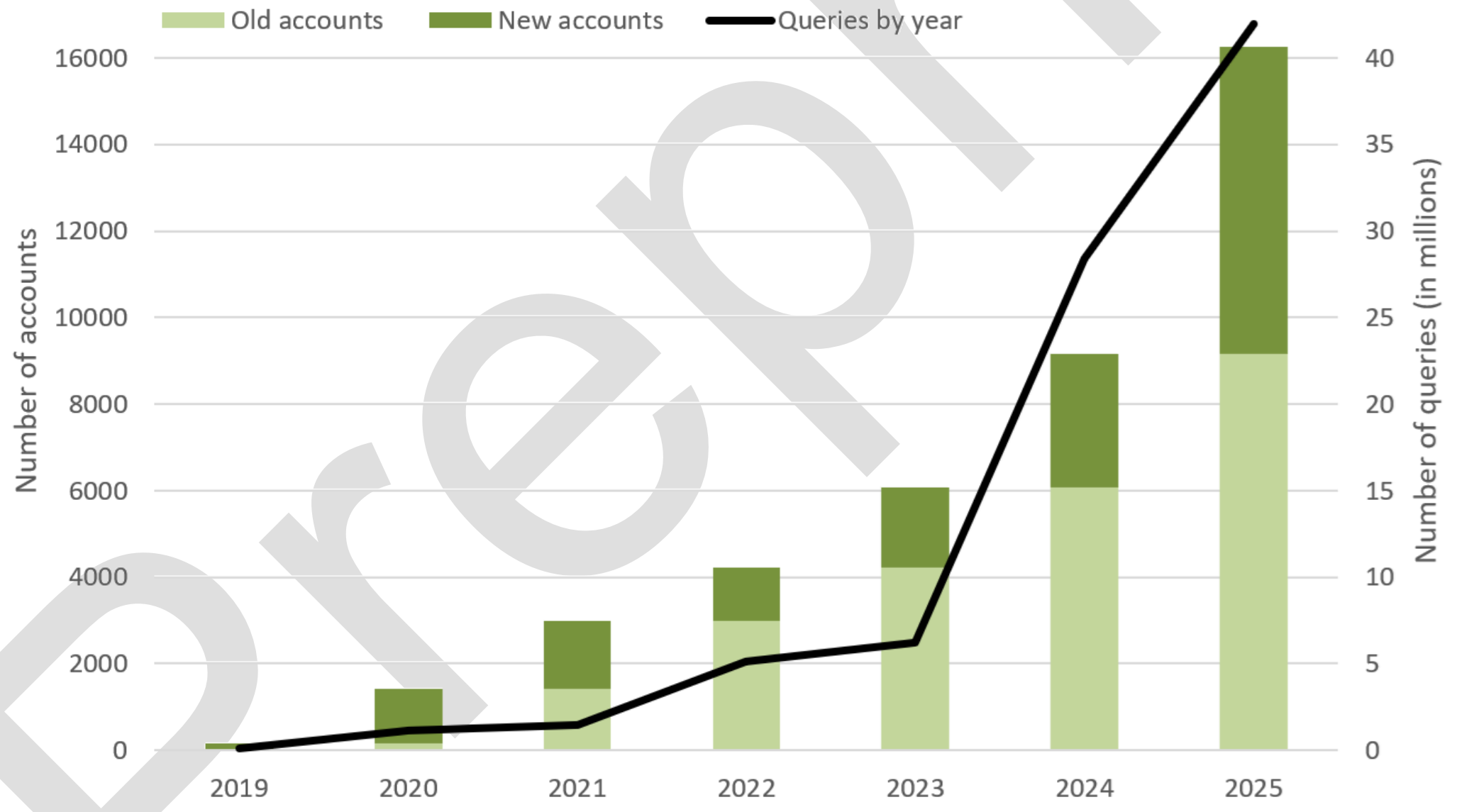

**Figure 1 :** Trends in the use of the Pl@ntNet API for machine-to-machine identification.

### *3.2. A genuine paradigm shift in ecological monitoring*

Pl@ntNet machine-to-machine service has led to tangible advances in ecology, resulting in scientific publications. Pernat et al. (2024), used it to identify the plants visible in a database of insect photographs in order to study the interactions between plants and pollinators. They developed a semi-automated data pipeline in which 1,741 images of *Isodontia mexicana* (a wasp) from iNaturalist were sent to Pl@ntNet to identify the plant species visited (**Figure 2**). This approach provided a better understanding of the plants visited by *Isodontia mexicana* in its native range and in the invasion area. Milenković et al., (2026) integrated the machine-to-machine service into their augmented-reality mobile citizen science app: Tree-Quest. This app enables users to identify and quantify the structural characteristics of individual trees. Volunteers using Tree-Quest were able to produce data of a quality very close to that of professional forest inventory surveyors. Finally, Ridding et al. (2026) developed the citizen science app e-Surveyor for conducting flora surveys and predicting habitat types. Regardless of their level of botanical expertise, most participants were able to accurately identify nearly half of the plant species present and determine the correct habitat using e-Surveyor. This makes e-Surveyor a useful tool for supporting biological surveys.

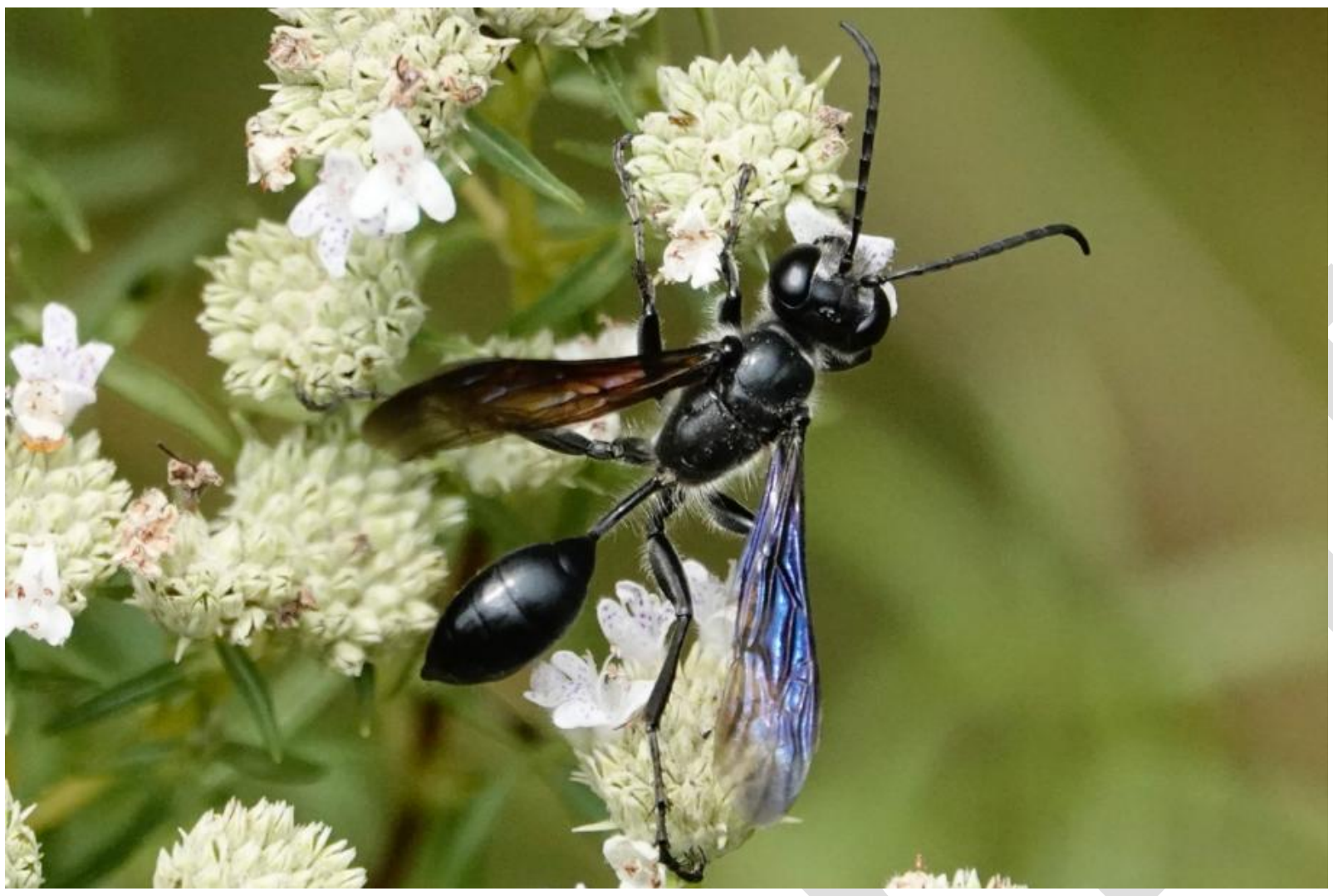

**Figure 2 :** A photograph of *Isodontia mexicana* published on iNaturalist (Pernat et al., 2024).

The machine-to-machine identification service has also paved the way for the processing of image collections that were not originally produced for ecological purposes. August et al., (2020) used this service to extract information on plant occurrence from more than 60,000 geotagged photographs shared on social media. This has made it possible to transform opportunistic image sources into valuable ecological datasets. For their part, Espitalier et al., (2025) and Goëau et al., (2026) used the machine-to-machine service to set up a workflow for estimating the presence and spread of invasive plants from images taken from vehicles. Espitalier et al., (2025) were able to carry out conclusive tests for the detection of six invasive plant taxa in Denmark, using 14,808 high-resolution images taken from vehicles. Such approaches are closely aligned with the growing use of drones, camera networks, and robotic platforms in ecological monitoring, where large numbers of images must be processed automatically to extract species-level information (Kissling et al., 2018).

Pl@ntNet machine-to-machine service also has applications in the agricultural sector. As part of the AGROSUS project (European Union “Horizon” research and innovation programme: grant no. 101084084), images captured automatically in a vineyard using an autonomous mobile robot were identified via the machine-to-machine service. This enabled the automatic and fast creation of a big image database of the plants present in the vineyard to facilitate the detection of weeds in vineyards (Emmi & Fernández Saavedra, 2025). Pl@ntNet machine-to-machine service is also a key element of the PARAD research project (project funded under the French government's PARSADA plan: “Anticipation of the withdrawal of active substances and development of alternative crop protection techniques”). It aims to develop selective weed control by utilising the machine-to-machine service to individually recognise plants from images captured by robots or agricultural machinery.

### 3.3. The wider implications

Pl@ntNet machine-to-machine service enables the implementation of private and public initiatives aimed at raising awareness and informing the general public about biodiversity issues. Although this deviates slightly from the original objective of biodiversity monitoring, it is an application that complements this objective and pursues the same goals. The public citizen science app e-Surveyor, described above, helps to strengthen users' trust in, knowledge of and engagement with nature (Ridding et al., 2026). There is also the free mobile app "Hidden Garden: Botanical Quest" (Minteract, 2026), developed for the Cephalonia Botanica botanical garden in Greece. It incorporates the Pl@ntNet machine-to-machine service to identify, in real time, the species photographed by visitors. It thus offers an interactive, fun and educational experience that encourages the discovery and appreciation of local flora.

## 4. Example

To illustrate the use of Pl@ntNet machine-to-machine service, we have drawn inspiration from the approach described by Pernat *et al.* (2024): the identification of plants in images of insects sourced from iNaturalist, with a view to studying the relationships between plants and pollinators (**Figure 2**). This workflow is available in the following Git repository: https://github.com/Beniofh/PlantNet_API_plant_pollinator. The workflow begins with the automatic extraction of metadata, including the URLs of the first images, from all 'scientific-quality' observations of an insect species on iNaturalist for a specific time period and geographical area. In this example, these are the observations of *Isodontia mexicana* recorded between 2007 and 2021 in Europe and North America: 2,424 observations. The first image from each observation is automatically loaded with its iNaturalist URL and sent to the Pl@ntNet machine-to-machine service with the Pl@ntNet API. For each observation, the Pl@ntNet API returns whether a plant has been detected. If so, the API returns a prediction of the visible organ (flower, leaf, whole plant, etc.) and a list of the five most likely plant species, along with their confidence scores. Using a confidence score threshold of 0.5 for classification as a 'plant with a flower', 1,633 observations were identified as having a flower. Of these 1,633 observations, only those with a confidence score greater than 0.5 for taxonomic identification were retained, amounting to a total of 708 observations of insects on flowers: **Figure 3**.

To evaluate the performance of the workflow, the supplementary data of Pernat et al. (2024) was used. These contain expert annotations for 1,694 of the 2,424 observations of *Isodontia mexicana* extracted using the workflow. These data indicate the presence or absence of the plant and specify whether a flower is visible. The scientific name of the plant is also provided for 936 of these 1,694 expert observations. This enabled us to estimate that our workflow can distinguish observations containing a flower from other observations with a macro-precision of 0.97 and a weighted precision for flowers of 0.98 (**Figure 4**). For observations classified as containing a flower, plant identification achieved a macro-accuracy with Bayesian smoothing of 0.88 for the genus and 0.92 for the family. In conclusion, within 30 minutes with a standard laptop, it was possible to generate a scientifically valid dataset comprising 708 observations of plant-pollinator interactions (**Figure 3**). This represents a major paradigm shift in the study of plant-pollinator interactions.

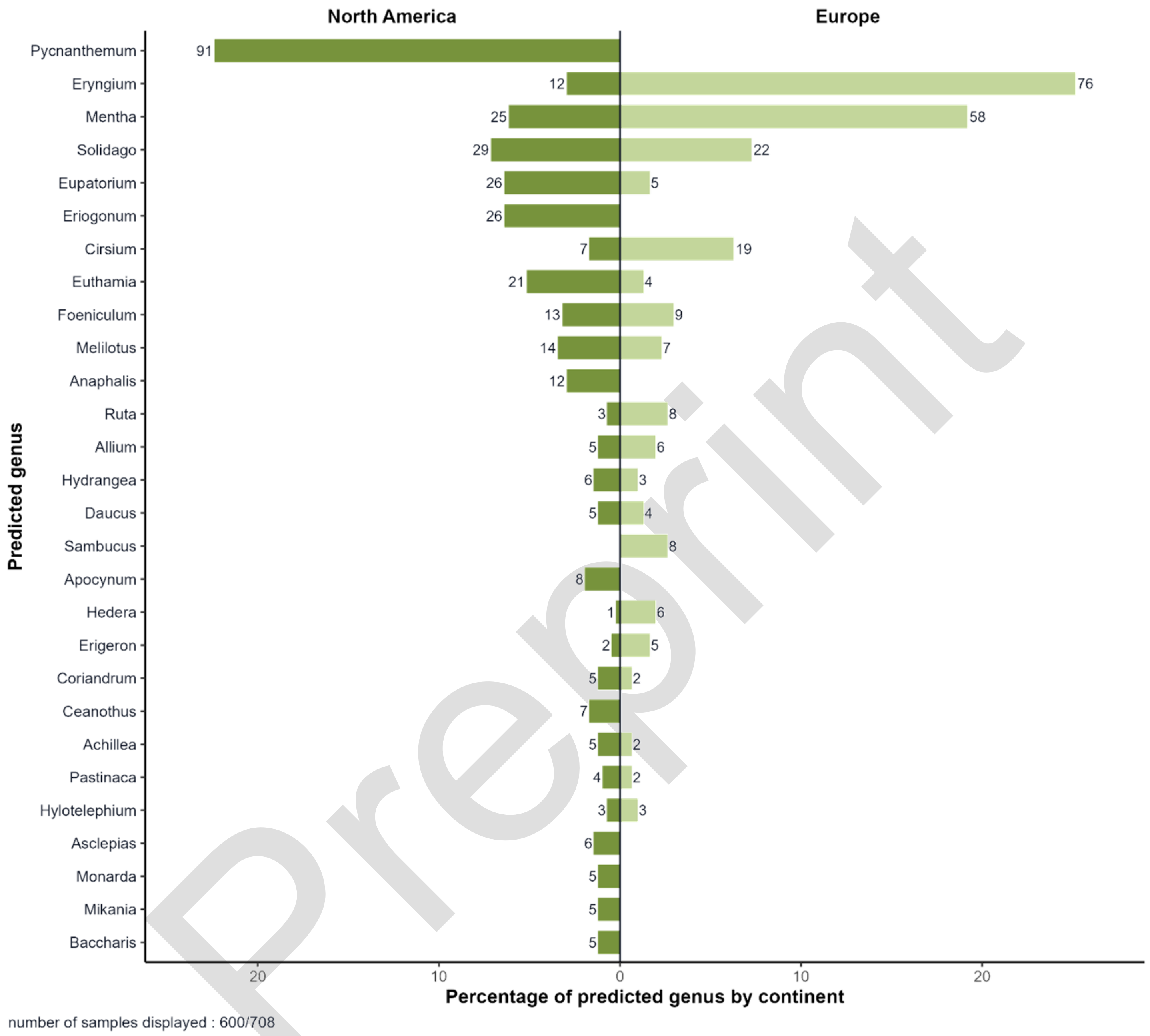

**Figure 3 :** Pl@ntNet predictions of flower genera (confidence score > 0.5) in images of Isodontia mexicana published on iNaturalist between 2007 and 2021. Only genera with ≥ 5 occurrences are shown. The values at the ends of the bars indicate the number of records per continent.

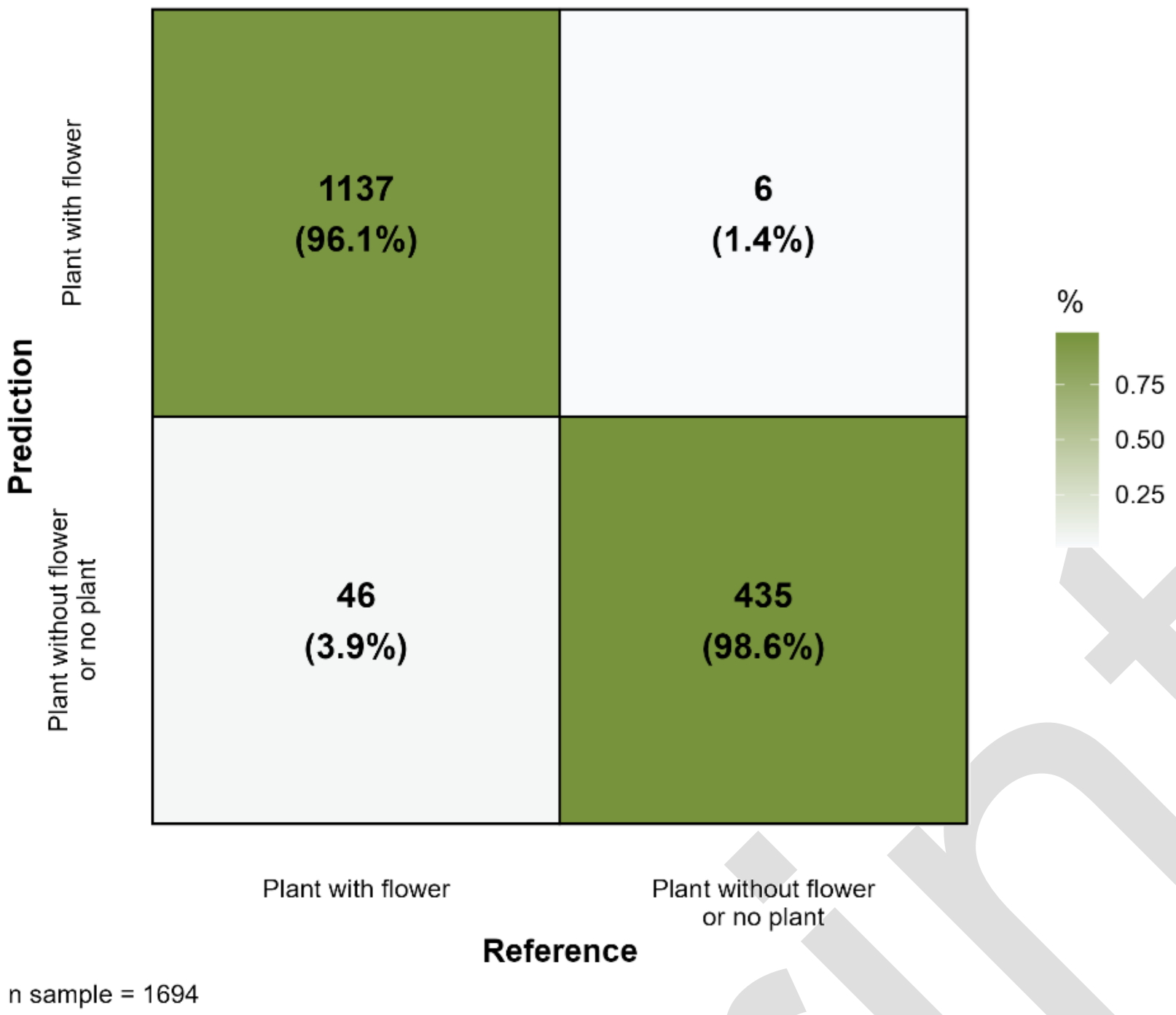

**Figure 4 :** Pl@ntNet's prediction of the presence of a flower in the image using a confidence score threshold of 0.5 for classification as a 'plant with a flower'.

## 5. Conclusion

The results presented in this article demonstrate that the Pl@ntNet machine-to-machine identification service is not only technically robust but also widely adopted, and that it is already producing tangible scientific and operational outcomes. Its growing use, with 40 million automated queries by 2025, confirms the need for reliable, robust identification services capable of operating without human intervention. This service marks a major advance in the development of large-scale, AI-based biodiversity monitoring infrastructure. By providing machine-to-machine access to large-scale plant identification models, this service overcomes the traditional limitations of user-driven mobile or web applications.

This machine-to-machine service enables the full integration of automated species recognition into the massive and heterogeneous ecological data streams generated by sensor networks and online platforms. Being 666 times faster than approaches requiring human intervention, it also makes real-time automated monitoring a serious possibility. By enabling seamless interaction between observation devices, databases and analytical workflows, this service helps to transcend current paradigms in ecology. It helps to bridge the gap between data acquisition and ecological understanding, which is an essential prerequisite for tackling the global biodiversity crisis.

## Acknowledgements

This work was funded by the French National Research Agency (ANR) through the grant Pl@ntAgroEco 22-PEAE0009. The work also received support from the European Commission through the GUARDEN (grant ID: 101060693) and the MAMBO (grant ID: 101060639) projects.

**Supplementary materials**

The supplementary material for this article (Tables S1 and S2) is available in a Zenodo repository at this address: https://doi.org/10.5281/zenodo.24112385 .

**Data and code**

The data required to reproduce the pre-processing carried out on the open-access datasets used in the article are available in the supplementary material. The R code and data required to reproduce the examples presented in the article are available on GitHub at the following address: https://github.com/Beniofh/PlantNet_API_plant_pollinator .